\documentclass[10pt,twocolumn,letterpaper]{article}

\usepackage[pagenumbers]{cvpr}

\usepackage{caption}

\DeclareMathOperator*{\argmax}{arg\,max}

\usepackage{multirow}
\usepackage{array}
\usepackage{bm}

\usepackage{colortbl}
\definecolor{oursbg}{HTML}{EDE6F7}
\definecolor{grpbg}{HTML}{F1F2F4}
\definecolor{refgray}{HTML}{7C828C}
\definecolor{dnc}{HTML}{B03A2E}
\definecolor{barA}{HTML}{C4C7CD}
\definecolor{barO}{HTML}{8A63C2}
\newcommand{\oursrow}{\rowcolor{oursbg}}
\newcommand{\grprow}[2]{\rowcolor{grpbg}\multicolumn{#1}{l}{\textit{#2}} \\}
\newcommand{\sbest}[1]{\underline{#1}}
\newcommand{\gr}[1]{\textcolor{refgray}{#1}}
\newcommand{\na}{\textcolor{refgray}{--}}
\newcommand{\dn}[1]{{\scriptsize\textcolor{dnc}{$\downarrow$\,#1}}}
\newcommand{\dl}[1]{{\scriptsize\textcolor{refgray}{#1}}}
\newcommand{\up}[1]{{\scriptsize\textcolor{refgray}{$\uparrow$\,#1}}}
\newcommand{\nb}[2][2.3em]{\makebox[#1][r]{#2}}

\makeatletter
\renewcommand\paragraph{\@startsection{paragraph}{4}{\z@}{0pt}{-0.5em}{\normalfont\normalsize\bfseries}}
\makeatother

\newcommand{\pullup}{\looseness=-1\relax}

\newcommand{\method}{RecallFlow\xspace}
\newcommand{\RR}{Recall--Relocalization\xspace}
\newcommand{\RA}{Recall Anchors\xspace}
\newcommand{\suite}{RecallSuite\xspace}

\newcommand{\taskPlate}{\textsc{MarkedPlate}\xspace}
\newcommand{\taskNoRep}{\textsc{NoRepeat}\xspace}
\newcommand{\taskNoRepFour}{\textsc{NoRepeat-4}\xspace}
\newcommand{\taskPair}{\textsc{MarkedPair}\xspace}
\newcommand{\taskCount}{\textsc{CountSwing}\xspace}

\newcommand{\Zt}{Z_t}
\newcommand{\Zhat}{\hat{Z}_t}
\newcommand{\Astar}{A^{\star}}
\newcommand{\N}{\mathcal{N}}
\newcommand{\Mem}{\mathcal{M}}
\newcommand{\Rec}{\mathcal{R}_t}
\newcommand{\sg}{\operatorname{sg}}
\newcommand{\nul}{\varnothing}

\usepackage{tikz}

\definecolor{cvprblue}{rgb}{0.21,0.49,0.74}
\usepackage[pagebackref,breaklinks,colorlinks,allcolors=cvprblue]{hyperref}

\ifdefined\pdfsuppressptexinfo \pdfsuppressptexinfo=14 \fi

\title{\method: Episodic Memory as the Source Distribution\\of Generative Robot Policies}

\author{%
Xuji Zhang \qquad Qilin Zhong \qquad Zichen Zhang \qquad Wei Zhai\\
Jian Di \qquad Kun Li \qquad Yang Cao \qquad Wenjun Lv$^{\dagger}$\\
University of Science and Technology of China\\
{\small $^{\dagger}$Corresponding author}\\
{\small Project page: \url{https://xujizhang.github.io/RecallFlow/}}
}

\begin{document}

\maketitle

\begin{abstract}
Diffusion and flow-matching policies have become a standard choice for visuomotor imitation learning, owing to their ability to model multimodal action distributions.
In memory-dependent manipulation, the robot must act on information that is no longer visible.
Existing methods feed the observation history to the policy as an additional condition, while action generation still starts from Gaussian noise.
However, the policy must then pick out the relevant target from a long and redundant history, leading to spurious correlations and unreliable decisions as the memory load grows.
Additionally, the initial noise carries no information about what the robot has seen, leaving generation without guidance toward the remembered target.
To address these issues, we propose \method, a flow-policy framework that builds the source distribution from episodic memory.
\RR relates the memory to the current view through a few recall slots, so that the source reflects where the remembered targets are now.
\RA then turns each slot into a Gaussian component of the source, enabling generation to start close to the intended action instead of from noise.
The transport network then completes the action from the current observation alone, keeping inference nearly as cheap as a memoryless policy.
We further build a suite of memory-dependent tasks grouped by memory load and compare all policies under a matched protocol.
Extensive experiments show that \method achieves a state-of-the-art average success rate of 80.4\% on the public MIKASA-Robo memory benchmark, and outperforms the strongest baseline by 20.3 points on our suite and by 18.8 points on four real-robot tasks.
\end{abstract}

\begin{figure}[t]
\centering
\includegraphics[width=\linewidth]{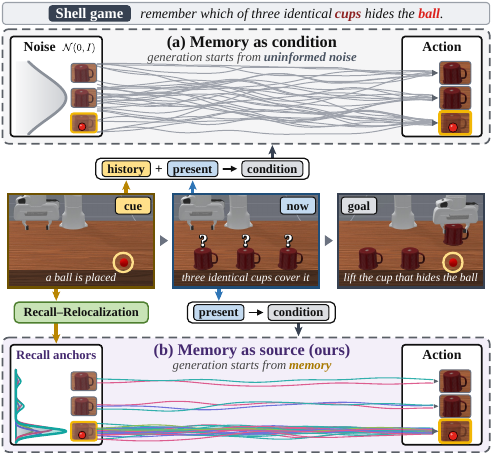}
\caption{\textbf{Start from what you remember.} In memory-dependent manipulation, the evidence that decides \emph{which} action to take (\emph{cue}) disappears before the robot acts (\emph{now}). (a) Feeding the memory to the policy as a condition still starts generation from uninformed noise. (b) \method starts generation from what it remembers: the episodic memory \emph{is} the source, and the transport network sees only the present.}
\label{fig:teaser}
\end{figure}

\section{Introduction}
\label{sec:intro}

Visuomotor policies learned from demonstrations now handle a wide range of manipulation tasks~\cite{act,dp,octo,openvla,pi0}.
They typically act only on what is currently in view.
Many manipulation tasks, however, depend on memory: the robot must act on information that is no longer visible.
Consider the shell game (\cref{fig:teaser}): a ball is placed on the table and then covered by one of three identical cups.
Once the cups are down, nothing in view sets that cup apart, yet the robot must still lift it.
Other tasks require remembering which containers have been touched or where an object was picked up~\cite{mikasa,gmp,chameleon}.
In such tasks, the current observation shows what \emph{can} be done but not what \emph{should} be done, leaving memoryless policies to guess.

Most visuomotor policies today are generative.
Diffusion and flow-matching policies~\cite{dp,flowmatching,rectflow,pi0} transport samples from a simple source distribution, usually an isotropic Gaussian, to the distribution of demonstrated actions, conditioned on the current observation.
To handle memory-dependent tasks, recent methods extend this condition with the past: they feed long observation windows to the policy~\cite{robomimic,ptp}, read the history through gated cross-attention~\cite{gmp}, or maintain structured memory states~\cite{chameleon,vpwem,memoryvla}.\pullup

Despite this progress, two issues remain.
First, memory enters only as an additional condition, so a single network must both pick out the relevant target from a long and redundant history and decide how to move.
This invites spurious correlations~\cite{causalconfusion,copycat} and can make such policies fall far below memoryless ones on tasks that need no memory~\cite{gmp}.
Second, generation still starts from uninformed noise, leaving it without guidance toward the remembered target.
Learned priors can replace this noise, but existing ones are derived from dataset statistics, the current observation, recent actions, other episodes, or a compressed history~\cite{diffusiondrive,leap,a2a,optimusvla,chronos}, not from the targets seen earlier in the same episode.
Together, these issues make history-conditioned policies increasingly unreliable as the memory load grows.\pullup

We take a different view and argue that the memory of an episode belongs in the source.
Flow matching can transport any source distribution to the action distribution~\cite{flowmatching,rectflow}, so the source is free to carry what the robot remembers.
Generation would then start near the remembered target, and the network would only need to refine the motion from the current observation.
However, memory cannot simply be plugged into the source.
What was seen earlier must first be found again in the current view, a problem long studied in video understanding~\cite{cater,permatrack,tcow,xmem,sam2}.
It must then be expressed as a distribution over actions.

\begin{figure*}[t]
\centering
\includegraphics[width=\linewidth]{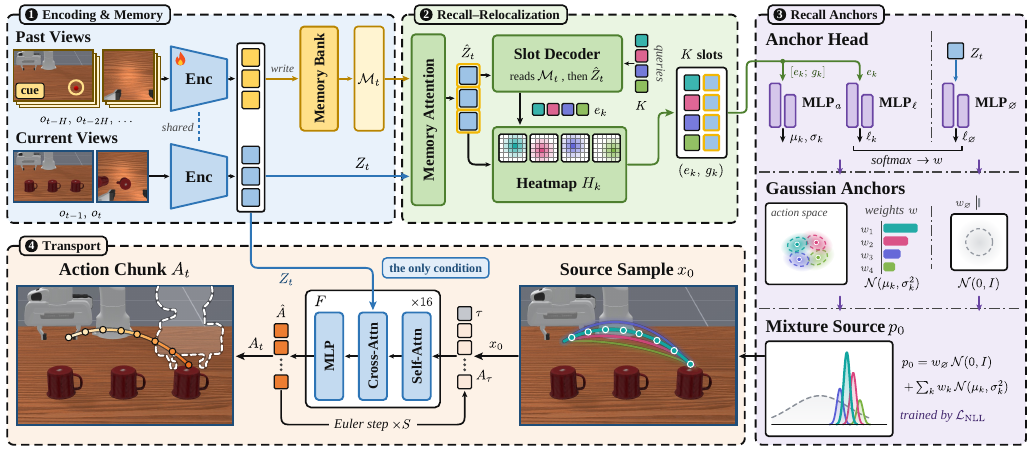}
\caption{\textbf{Overview of \method.}
The present observation is encoded into tokens $\Zt$.
\RR lets $\Zt$ read the episodic memory bank and decodes $K$ recall slots, each attending to the current view through a heatmap $H_k$.
\RA turns each slot into a Gaussian anchor in action space; with a standard Gaussian, the anchors form the source $p_0$.
A transport network conditioned only on $\Zt$ carries $x_0 \sim p_0$ to the action chunk, so memory reaches the action solely through the source.}
\label{fig:overview}
\end{figure*}

To this end, we propose \method, a flow-policy framework that builds the source distribution from episodic memory (\cref{fig:overview}).
It rests on a division of labor: memory decides \emph{where generation starts}, and the present decides \emph{how to get there}.\pullup

\paragraph{\RR: relating memory to the current view.}
\method keeps an episodic memory bank of patch-level tokens from earlier frames.
Memory attention lets the current visual tokens read this bank.
A few learnable recall slots then read both the memory and the current view, so that the source reflects where the remembered targets are now.\pullup

\paragraph{\RA: starting generation from memory.}
Each recall slot is mapped to a Gaussian anchor in action space, and the anchors, together with a standard Gaussian, form a mixture source.
Generation thus starts close to the intended action instead of from noise.
A transport network then completes the action chunk from the current observation alone, keeping inference nearly as cheap as a memoryless policy.

On \suite, our suite of five memory-dependent tasks, \method improves the average success rate by 20.3 points over the strongest of two memoryless and two history-conditioned policies under the same protocol, with the largest margins where several items must be remembered (64.7\% \vs 15.0\% on \taskNoRepFour).
On the public MIKASA-Robo benchmark~\cite{mikasa}, it reaches 80.4\% average success over five tasks, above the best published result of 75.1\%~\cite{chameleon}.
On RoboMimic~\cite{robomimic}, where the current observation suffices, it stays on par with memoryless policies, whereas long-history policies degrade.
On a real xArm~7, it improves over the strongest baseline by 18.8 points on four memory-dependent tasks.
Ablations show that forming the source from the episodic memory matches or exceeds conditioning the same transport network on it (97.0\% \vs 92.0\% on \taskNoRep), and that relocalization matters most when the memory load is high.
Our contributions are as follows:
\begin{itemize}[leftmargin=*,topsep=2pt,itemsep=1pt]
\item \textbf{Memory as source.} We form the source distribution of a flow policy from episodic memory as likelihood-trained Gaussian anchors, so that generation starts close to the intended action.
\item \textbf{\RR.} Memory attention and a few recall slots relate the episodic memory to the current view, so that the anchors reflect where the remembered targets are now.
\item \textbf{Evaluation.} We build a suite of five memory-dependent tasks grouped by memory load. Under a matched protocol, \method achieves the highest average success rate on it, and it sets the state of the art on the public MIKASA-Robo benchmark.
\end{itemize}

\section{Related Work}
\label{sec:related}

\paragraph{Memory-dependent manipulation.}
Visuomotor policies usually condition on a single frame or a short window of observations~\cite{dp,act}.
The simplest form of memory extends this window to a long history~\cite{simplememvla} or the whole episode, which raises the training cost and invites spurious correlations~\cite{causalconfusion,copycat,bpp}.
To make long histories usable, one line of work regularizes or compresses them~\cite{ptp,gmp}.
Another keeps only a few key frames selected by a vision--language model~\cite{bpp,memer}.
A third line maintains structured memory states: an event-level memory indexed by the control state~\cite{chameleon}, a slot memory addressed by the robot's trajectory~\cite{trace}, a combination of working and episodic memory~\cite{vpwem}, or memory modules inside vision--language--action models~\cite{memoryvla,mem}.
SAM2Act~\cite{sam2act} brings the memory of a video segmentation model to keyframe-based manipulation, and HAVE~\cite{have} ranks the samples of a memoryless generator with a history-aware verifier.
However, they all use memory as a condition or as a selector of samples, while generation still starts from uninformed noise.
In contrast, \method delivers memory only through the source distribution.

\paragraph{Non-Gaussian sources and learned priors.}
Flow matching and rectified flow~\cite{flowmatching,rectflow} can transport any source distribution, including sources coupled to the target~\cite{datacoupling}.
Beyond robotics, generation has been started from another modality, from data-dependent priors, or from optimized, predicted, or warped noise~\cite{crossflow,priorgrad,led,initno,goldennoise,gowiththeflow}.
In driving and robotics, DiffusionDrive~\cite{diffusiondrive} starts from anchored Gaussians clustered over the dataset, which cuts the number of denoising steps.
Other methods derive the source from the current observation~\cite{leap,bridgepolicy,vita,resvla}, from informative prior policies~\cite{bridger,navibridger}, or from previous actions~\cite{a2a,temporalpolicy,warmprior,streamingflow}, and a related line steers frozen policies in their noise space~\cite{dsrl,frs,flowdagger}.
Closest to ours, OptimusVLA~\cite{optimusvla} averages trajectories retrieved from other episodes into a prior, Chronos~\cite{chronos} generates a coarse prior from a compressed history, SL-FM~\cite{slfm} learns a state-conditioned Gaussian-mixture source by likelihood, and MOSS~\cite{moss} builds mode-specific Gaussian sources so that the starting point selects the behavior mode.
However, none of these sources keeps track of which target was seen earlier in the episode or where it is now.
\method also learns a mixture source by likelihood, but forms it from episodic memory relocalized in the current view.

\paragraph{Memory attention in video understanding.}
Reasoning about objects that are no longer visible has long been studied in video understanding~\cite{cater,permatrack,tcow}.
Space-time memory networks~\cite{stm} segment objects by letting the current frame attend to a memory of past frames.
XMem~\cite{xmem} organizes this memory for long videos, Cutie~\cite{cutie} reads it at the object level, and SAM~2~\cite{sam2} makes memory attention the core of promptable video segmentation.
Long-video understanding compresses memory banks by merging similar entries~\cite{malmm}, and set prediction reads one object per learnable query~\cite{detr,slotattention}.
\RR builds on memory attention and learnable queries, but its output is not a mask: it defines where action generation starts.

\section{Method}
\label{sec:method}

As shown in \cref{fig:overview}, \method adds two components to a flow policy.
\RR (RR) relates the episodic memory to the current view, and \RA (RA) turns the result into the source distribution, so that generation starts from what the robot remembers.
Formally, with $o_t$ the camera images at step $t$ and $\Mem_t$ the episodic memory of the current episode, an action chunk $A_t$ is generated as
\begin{align}
\Zt &= \mathrm{Enc}(o_{t-1}, o_t), \\
\Rec &= \mathrm{RR}(\Zt, \Mem_t), \\
p_0 &= \mathrm{RA}(\Rec, \Zt), \\
A_t &= \mathrm{Transport}(x_0 \sim p_0 \mid \Zt),
\end{align}
where $\Rec = \{(e_k, g_k)\}_{k=1}^{K}$ collects the outputs of the $K$ recall slots.
Memory thus reaches the action only through the source $p_0$, while the transport network sees only the present tokens $\Zt$.

\subsection{Preliminaries}
\label{sec:prelim}

\paragraph{Flow policies.}
A visuomotor policy predicts an action chunk $A \in \mathbb{R}^{H \times d}$ of $H$ consecutive $d$-dimensional actions.
Flow policies model the chunk with rectified flow~\cite{rectflow,flowmatching}, which transports samples of a source distribution $p_0$ to demonstrated chunks along straight paths.
Given a source sample $x_0 \sim p_0$ and a demonstrated chunk $\Astar$, a network $F$ receives the interpolant
\begin{equation}
A_\tau = (1 - \tau)\, x_0 + \tau\, \Astar, \qquad \tau \in [0, 1),
\label{eq:interp}
\end{equation}
and predicts the clean chunk $\hat{A} = F(A_\tau, \tau, \Zt)$; it is trained with $\lVert \hat{A} - \Astar \rVert^2$.
At inference, the velocity $v = (\hat{A} - A_\tau)/(1 - \tau)$ is integrated from $x_0$ with $S$ Euler steps.
The source is almost always $\N(0, I)$, as in denoising diffusion~\cite{ddpm,ddim}.
However, the same objective holds for any $p_0$, so \method only needs to change $p_0$.

\paragraph{Present tokens and transport network.}
An image encoder maps the third-person and wrist images of each control step to grids of patch tokens, average-pooled to $8 \times 8$.
The present tokens $\Zt$ collect the tokens of the current and the previous step from both cameras, which makes motion visible.
$\Zt$ is the only condition of the transport network $F$.
The encoder is SigLIP~2~\cite{siglip2} and the transport network is a DiT~\cite{dit}; our implementation builds on the public code of GMP~\cite{gmp}.
Without the two components introduced below, the policy is a memoryless flow policy, denoted FM, which shares the encoder, transport network, and training recipe with \method and serves as its memoryless counterpart.

\subsection{\RR}
\label{sec:rr}

What the robot remembers was seen in the past, but the action must be generated for the present.
By then the arm has moved, the wrist camera with it, and the cue itself may be gone.
A memory can therefore guide generation only after it has been related to the current view.
Motivated by this, \RR proceeds in three steps (\cref{fig:overview}, center): the present tokens read an episodic memory bank, a few recall slots read the memory and then the present, and each slot pools the present-view evidence for what it recalls.

\paragraph{Episodic memory bank.}
\label{sec:memory}
After each executed chunk, \method writes the patch tokens of the frame at which the chunk started, from both cameras, into a memory bank $\Mem$.
Keeping the tokens at patch level preserves the spatial detail needed to relate a memory to the current view.
Each token is tagged with learned embeddings of its camera, its grid position, and its age measured in chunks.
An entry thus corresponds to one real moment of the episode.
Entries are merged only when the bank is full: with a capacity of $C$ entries, the two most similar adjacent entries are merged position by position.
At the first chunk of an episode the memory is empty, and all recall slots are masked.

\paragraph{Memory attention.}
The present tokens first read the memory, in the same way as the current frame reads past frames in video object segmentation~\cite{stm,sam2}.
Memory attention has $L_m$ layers, each applying self-attention over the present tokens, cross-attention from the present tokens to $\Mem$, and an MLP.
It outputs memory-conditioned present tokens $\Zhat$.

\paragraph{Recall slots.}
$K$ learnable queries, the recall slots, are decoded by $L_q$ layers.
In each layer, the slots attend to each other, then to $\Mem$ (what is remembered), and then to $\Zhat$ (where it is now).
The output is one embedding $e_k$ per slot, which turns a memory of variable size into a fixed set of candidates for the source.

\paragraph{Recall heatmaps.}
Each slot then attends to the current view through a heatmap:
\begin{align}
H_k(p) &= \operatorname*{softmax}_{p} \left( \frac{\langle e_k, \Zhat(p) \rangle}{\sqrt{D}} \right), \label{eq:heatmap} \\
g_k &= \sum_{p} H_k(p)\, \Zhat(p), \label{eq:pool}
\end{align}
where $p$ ranges over the current-frame patches of one camera and $D$ is the token dimension.
Each slot thus has one heatmap per camera, and $g_k$ concatenates the pooled features of the two cameras.
$H_k$ is a distribution over the image, and $g_k$ is the present-view feature pooled with it.

The outputs $(e_k, g_k)$ go only to \RA.
Neither the memory nor the slots reach the transport network, so memory enters the action through a single channel and the transport network stays as light as a memoryless one.\pullup

\subsection{\RA}
\label{sec:ra}

A standard flow policy starts generation from $\N(0, I)$, which carries no information about what the robot has seen.
Motivated by this, \RA forms the source from the recall slots (\cref{fig:overview}, right): each slot becomes a Gaussian anchor in action space.
This moves the starting point of generation from noise to the neighborhood of the intended action, leaving the transport network only a refinement.

\paragraph{Anchors.}
Each slot is mapped to the mean and standard deviation of one diagonal Gaussian over the action chunk,
\begin{equation}
(\mu_k, \log \sigma_k) = \mathrm{MLP}_a([e_k; g_k]), \quad \mu_k, \sigma_k \in \mathbb{R}^{H \times d},
\label{eq:anchor}
\end{equation}
with $\sigma_k$ bounded below by $\sigma_{\min}$.

\paragraph{Source distribution.}
Each slot receives a logit $\ell_k = \mathrm{MLP}_\ell(e_k)$, and a no-recall logit $\ell_\nul = \mathrm{MLP}_\nul(\operatorname{pool}(\Zt))$ is computed from the present alone.
With the weights $w = \operatorname{softmax}([\ell_\nul, \ell_1, \dots, \ell_K])$, the source is
\begin{equation}
p_0 = w_\nul\, \N(0, I) + \sum_{k=1}^{K} w_k\, \N(\mu_k, \sigma_k^2).
\label{eq:source}
\end{equation}
The no-recall component lets the policy start from standard noise when it has nothing to recall.
While the memory is empty, all slots are masked and the source is exactly $\N(0, I)$.
More generally, removing the memory reduces \method to the architecture and sampling procedure of the memoryless flow policy FM.

\paragraph{Learning the anchors.}
An anchor is useful only if it lies close to the action that should be taken.
We therefore train the source by maximum likelihood: we minimize the negative log-likelihood of the demonstrated chunk $\Astar$ under the mixture,\pullup
\begin{equation}
\mathcal{L}_{\mathrm{NLL}} = -\frac{1}{Hd} \log p_0(\Astar).
\label{eq:nll}
\end{equation}
This loss moves the anchors towards the demonstrated action and raises the weight of the anchor that explains it best.
Since the anchors are computed from the recall slots, the same loss also trains \RR.

We do not train the anchors with the flow loss that trains the transport network (\cref{sec:train}), because it rewards the wrong property.
It asks the network to recover $\Astar$ from an interpolant between a source sample and $\Astar$.
When the source is narrow, the source sample is almost known in advance, so the interpolant gives $\Astar$ away even if the source carries no information about it.
The flow loss would thus reward anchors for being narrow rather than for being right.

Two regularizers complete the objective: a KL term $\mathcal{L}_{\mathrm{KL}}$ pulls each anchor to $\N(0, I)$, and a relaxed winner-takes-all term $\mathcal{L}_{\varepsilon}$~\cite{multihyp} keeps every slot trained (\cref{sec:supp-impl}).

\subsection{Training and Inference}
\label{sec:train}

\paragraph{Training.}
A training sample is a window of an episode.
The memory bank is rolled out chunk by chunk inside the window, exactly as at inference.
The transport network is trained on source samples drawn from the component that explains the demonstration.
This data-dependent coupling~\cite{datacoupling} matches sampling at test time when the weights $w$ are calibrated.
Concretely, a component $c$ is sampled from the posterior $r_c \propto w_c\, \N(\Astar; \mu_c, \sigma_c^2)$, the source sample is $x_0 \sim \N(\sg(\mu_c), \sg(\sigma_c)^2)$, and the flow loss is
\begin{equation}
\mathcal{L}_{\mathrm{flow}} = \big\lVert F(A_\tau, \tau, \Zt) - \Astar \big\rVert^2,
\label{eq:flow}
\end{equation}
with $A_\tau$ from \cref{eq:interp} and $\tau \sim \mathcal{U}[0, 1)$.
The stop-gradient $\sg(\cdot)$ keeps the flow loss from reaching the anchors.
The overall training objective is
\begin{equation}
\mathcal{L} = \mathcal{L}_{\mathrm{NLL}} + \lambda\, \mathcal{L}_{\mathrm{flow}} + \beta\, \mathcal{L}_{\mathrm{KL}} + \mathcal{L}_{\varepsilon}.
\label{eq:total}
\end{equation}

\paragraph{Inference.}
For each action chunk, \method encodes the present, runs \RR and \RA, picks a source sample, integrates $S$ Euler steps, executes the chunk, and writes the chunk's first frame into the memory.
For evaluation we use a deterministic mode: the policy recalls if the total recall weight $\sum_k w_k$ exceeds the no-recall weight $w_\nul$ and then starts from the mean of the highest-weighted anchor; otherwise it starts from $0$.
Comparing the totals first prevents several slots that share one target from splitting its weight below $w_\nul$.
\cref{sec:supp-impl} gives the stochastic mode and further details.

\section{Experiments}
\label{sec:exp}

\begin{figure*}[t]
\centering
\includegraphics[width=\linewidth]{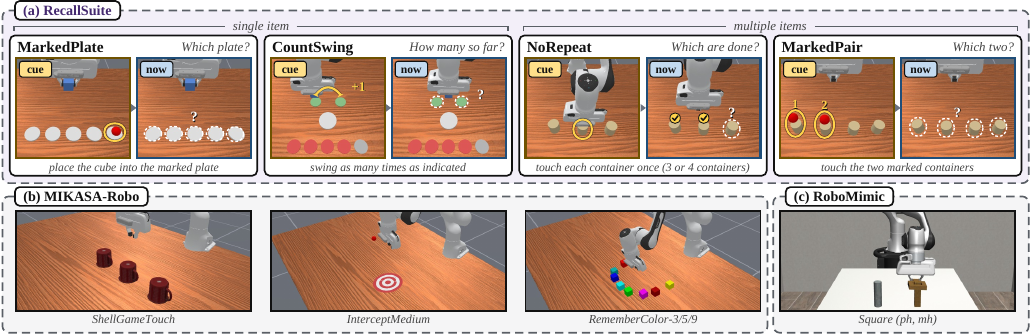}
\caption{\textbf{Evaluation tasks in simulation.} (a) \suite consists of five memory-dependent tasks: the cue is seen only earlier in the episode, and the candidates look identical when the robot must decide. (b), (c) Tasks from MIKASA-Robo and RoboMimic.}
\label{fig:tasks}
\end{figure*}

\begin{table*}[t]
\centering
\caption{\textbf{Comparison on \suite.} We report the success rate (\%) averaged over three training seeds. \method is on par with long-history DP when one item must be remembered and outperforms all baselines when several must be.}
\label{tab:suite}
\small
\setlength{\tabcolsep}{8.5pt}
\begin{tabular}{lcccccc}
\toprule
& \multicolumn{2}{c}{\textbf{Single item}} & \multicolumn{3}{c}{\textbf{Multiple items}} & \\
\cmidrule(lr){2-3}\cmidrule(lr){4-6}
Method & \taskPlate & \taskCount & \taskNoRep & \taskNoRepFour & \taskPair & \textbf{Avg.}\,$\uparrow$ \\
\midrule
\gr{\textit{Chance}} & \gr{20.0} & \gr{11.7} & \gr{22.2} & \gr{9.4} & \gr{16.7} & \gr{16.0} \\
\midrule
\grprow{7}{Memoryless}
DP~\cite{dp} & 17.3 & 10.0 & 1.3 & 0.0 & 1.3 & 6.0 \\
FM & 16.0 & 6.3 & 13.3 & 2.3 & 0.0 & 7.6 \\
\midrule
\grprow{7}{History-conditioned}
Medium-history DP & \textbf{100.0} & 10.3 & 31.7 & 1.7 & 4.7 & 29.7 \\
Long-history DP & \textbf{100.0} & \textbf{99.0} & \sbest{70.7} & \sbest{15.0} & \sbest{36.7} & \sbest{64.3} \\
\midrule
\grprow{7}{Memory as source}
\oursrow \textbf{\method} & \sbest{95.7} & \sbest{98.3} & \textbf{97.0} & \textbf{64.7} & \textbf{67.3} & \textbf{84.6} \\
\bottomrule
\end{tabular}
\end{table*}

\subsection{Experimental Setup}
\label{sec:exp-setup}

\paragraph{Benchmarks.}
We evaluate on three simulated benchmarks and a real robot, each answering one question.
\emph{\suite}, our suite of five memory-dependent tasks (\cref{fig:tasks}), tests how policies cope with a growing memory load and hosts all protocol-matched comparisons and ablations.
\emph{MIKASA-Robo}~\cite{mikasa} is a public memory benchmark, on which we compare with published methods; we use its five memory-intensive tasks with 250 demonstrations each, the setting in which these methods report their results~\cite{gmp,memoryvla,chameleon}.
\emph{RoboMimic}~\cite{robomimic} contains tasks that the current observation suffices to solve, on which we test whether memory hurts when it is not needed.
The real-robot experiments are described in \cref{sec:exp-real}.

\paragraph{Tasks of \suite.}
The five tasks fall into two groups of memory load.
In the single-item tasks, one remembered quantity decides the action.
In \taskPlate, a marker descends into one of five identical plates and leaves, and the robot must place a cube into that plate.
In \taskCount, the robot must swing a cube between two pads as many times as a row of indicators shows, which requires remembering how many swings it has made.
In the multi-item tasks, the robot must keep several events in mind.
In \taskNoRep, it must touch each of three identical containers exactly once, and \taskNoRepFour uses four containers.
In \taskPair, a marker hovers over two of four identical containers in turn, and the robot must then touch both.
Each task comes with 250 scripted demonstrations; \cref{sec:supp-suite} gives the details and relates the tasks to those of prior work~\cite{mikasa,robomme,chameleon}.
We report the success rate, where a trial succeeds only if the first choice is correct and the task is completed, so that trial and error is not rewarded.

\paragraph{Baselines.}
We compare with two groups of policies.
The memoryless group contains Diffusion Policy (DP)~\cite{dp} and FM, the memoryless flow policy of \cref{sec:prelim}, \ie, \method without its memory.
The history-conditioned group contains medium- and long-history DP, which condition DP on the frames of the past 16 and 120 steps and thus represent memory used as a condition.
All baselines keep their released hyperparameters (\cref{sec:supp-baselines}).
Published memory methods are compared on MIKASA-Robo (\cref{tab:mikasa}).\pullup

\paragraph{Protocol.}
Within a benchmark, all policies that we train share the demonstrations and, on each task, the number of gradient steps.
We save checkpoints periodically, evaluate each on the same 100 episodes in deterministic mode, and report the best one for every policy.
On \suite and RoboMimic, we train three seeds per method and report the mean (\cref{sec:supp-results}).
Unless stated otherwise, policies observe a third-person and a wrist RGB image at $128 \times 128$, without proprioception, and output absolute joint targets.

\begin{table}[t]
\centering
\caption{\textbf{Comparison with published methods on MIKASA-Robo~\cite{mikasa}.} We report the success rate (\%) on ShellGameTouch (SGT), InterceptMedium (IM), and RememberColor with 3, 5, and 9 colors (RC3--RC9). \method, averaged over three seeds, achieves the highest average ($80.4 \pm 1.1$). $^\ast$Available only on arXiv. $^\ddagger$OpenVLA-OFT in~\cite{memoryvla}.}
\label{tab:mikasa}
\small
\setlength{\tabcolsep}{2.9pt}
\begin{tabular}{lcccccc}
\toprule
Method & SGT & IM & RC3 & RC5 & RC9 & \textbf{Avg.}\,$\uparrow$ \\
\midrule
Octo-small~\cite{octo,mikasa} & 46 & 39 & 45 & 17 & 11 & 31.6 \\
OpenVLA$^\ddagger$~\cite{openvla,mikasa} & 47 & 14 & 59 & 16 & 6 & 28.4 \\
$\pi_0$~\cite{pi0,mikasa} & 33 & 42 & 35 & 22 & 15 & 29.4 \\
SpatialVLA~\cite{spatialvla,mikasa} & 23 & 27 & 27 & 17 & 11 & 21.0 \\
CronusVLA~\cite{cronusvla,memoryvla} & 32 & 5 & 31 & 13 & 9 & 18.0 \\
MemoryVLA~\cite{memoryvla} & 88 & 24 & 44 & 30 & 20 & 41.2 \\
GMP~\cite{gmp} & 98 & \sbest{83} & 80 & 61 & 17 & 67.8 \\
SimpleMemVLA$^\ast$~\cite{simplememvla} & \sbest{99} & \sbest{83} & 71 & 58 & \textbf{59} & 74.0 \\
Chameleon~\cite{chameleon} & 96.7 & 58.9 & \sbest{94.4} & \sbest{72.2} & \sbest{53.3} & \sbest{75.1} \\
\oursrow \textbf{\method} & \textbf{100.0} & \textbf{87.3} & \textbf{100.0} & \textbf{78.3} & 36.3 & \textbf{80.4} \\
\bottomrule
\end{tabular}
\end{table}

\begin{figure*}[t]
\centering
\includegraphics[width=\linewidth]{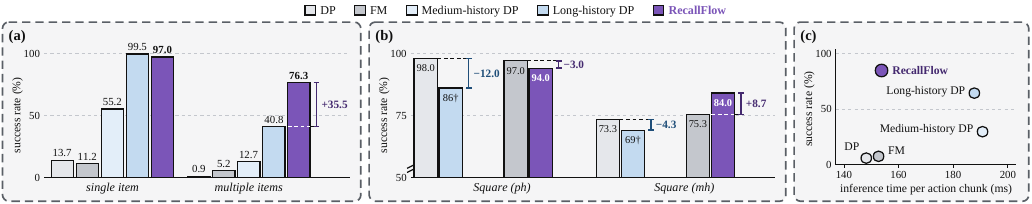}
\caption{\textbf{Effect of memory load, tasks without memory, and inference cost.} (a) On \suite, \method matches long-history DP when one item must be remembered and stays ahead of all baselines when several must be. (b) On RoboMimic, where memory is not needed, adding a long history lowers DP, whereas \method stays on par with its memoryless counterpart FM. (c) \method reaches the highest success rate on \suite at nearly the inference cost of a memoryless policy. $^\dagger$Reported in~\cite{gmp}.}
\label{fig:summary}
\end{figure*}

\subsection{Simulation Results}
\label{sec:exp-sim}

\paragraph{Results on \suite.}
\label{sec:exp-suite}
\cref{tab:suite} reports the results, and \cref{fig:summary}a summarizes them by memory load.
Memoryless policies never exceed chance, confirming that the current view does not determine the answer.
On the single-item tasks, a long history is sufficient: long-history DP reaches 100.0\% and 99.0\%, and \method is comparable (95.7\% and 98.3\%).
On the multi-item tasks, long-history DP falls to 70.7\% on \taskNoRep and to 15.0\% once a fourth container is added (\taskNoRepFour), and reaches 36.7\% on \taskPair, whereas \method reaches 97.0\%, 64.7\%, and 67.3\%.
On average, \method outperforms the strongest baseline by 20.3 points.

\paragraph{Comparison on MIKASA-Robo.}
\label{sec:exp-mikasa}
We train one policy per task (\cref{sec:supp-mikasa}); the numbers of the other methods are those reported in their papers.
As shown in \cref{tab:mikasa}, \method achieves the highest average success rate, 80.4\%, compared with 75.1\% for Chameleon~\cite{chameleon} and 67.8\% for GMP~\cite{gmp}.
It is the best on four of the five tasks.
Several methods are evaluated only on ShellGameTouch and RememberColor3~\cite{vpwem}; on these two tasks, \method reaches 100.0\% (\cref{sec:supp-mikasa}).

\paragraph{When memory is not needed: RoboMimic.}
\label{sec:exp-robomimic}
A policy with memory should not pay for it when the current observation already determines the action.
We test this on the Square task of RoboMimic with proficient (ph) and multi-human (mh) demonstrations (\cref{fig:summary}b; setup and numbers in \cref{sec:supp-robomimic}).
Long-history DP falls below the memoryless DP in both settings.
\method does not: it stays on par with the memoryless policies, 89.0\% on average against 86.2\% for FM and 85.7\% for DP, and is the best on the multi-human data.

\begin{figure*}[t]
\centering
\includegraphics[width=\linewidth]{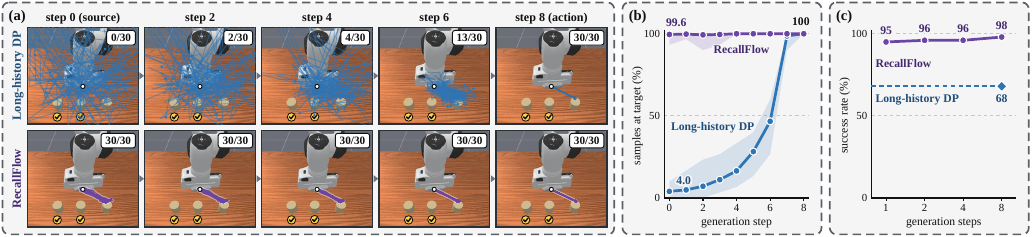}
\caption{\textbf{Memory decides where generation starts.} (a) At a decision of a \taskNoRep episode, all thirty samples of \method head to the untouched container from the source on, whereas long-history DP starts from noise and commits only in its last generation steps. (b) Over 15 decisions, the source samples of \method are already at the target. (c) \method therefore keeps its success rate with a single generation step.}
\label{fig:qual}
\end{figure*}

\begin{figure*}[t]
\centering
\includegraphics[width=\linewidth]{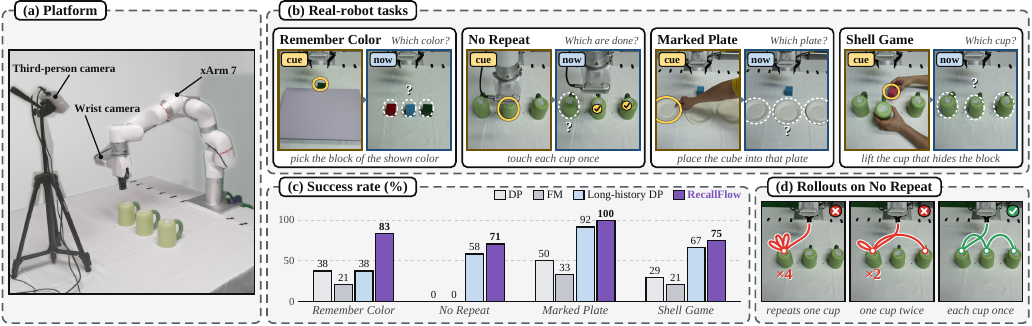}
\caption{\textbf{Real-robot experiments.} (a) We deploy the policies on an xArm~7 with a wrist camera and a third-person camera. (b) We design four tasks that differ in what must be remembered. (c) \method achieves the highest success rate on all four tasks. (d) The three behaviors observed on No Repeat: repeating one cup, touching one cup twice, and touching each cup once.}
\label{fig:real}
\end{figure*}

\subsection{Ablation Study}
\label{sec:exp-ablation}

We ablate three design choices on \taskPlate, \taskNoRep, and \taskPair and add one reference row; results on \taskCount are in \cref{sec:supp-ablation}.
\textbf{W/o memory} is FM.
\textbf{As condition} uses the memory in the common way: every memory entry is pooled into one token per camera and added to the condition of the transport network, and generation starts from $\N(0, I)$.
\textbf{W/o relocalization} removes memory attention and the slots' read of the present, so the anchors are computed from what is remembered alone.
\textbf{Flow loss only} removes the likelihood and KL terms and samples components with a straight-through Gumbel-softmax, so that the flow loss trains the anchors.

\begin{table}[t]
\centering
\caption{\textbf{Ablation study on \suite.} We report the success rate (\%), averaged over three seeds, and its change when one design choice of \method is altered.}
\label{tab:ablation}
\small
\setlength{\tabcolsep}{1.9pt}
\begin{tabular}{@{}llll@{}}
\toprule
Variant & \taskPlate & \taskNoRep & \taskPair \\
\midrule
\oursrow \textbf{\method} & \nb{\textbf{95.7}} & \nb{\textbf{97.0}} & \nb{\textbf{67.3}} \\
\midrule
W/o memory & \nb{16.0}\,\dn{79.7} & \nb{13.3}\,\dn{83.7} & \nb{0.0}\,\dn{67.3} \\
As condition & \nb{94.7}\,\dn{1.0} & \nb{92.0}\,\dn{5.0} & \nb{51.0}\,\dn{16.3} \\
W/o relocalization & \nb{98.3}\,\up{2.6} & \nb{84.7}\,\dn{12.3} & \nb{56.3}\,\dn{11.0} \\
Flow loss only & \nb{53.3}\,\dn{42.4} & \nb{98.0}\,\up{1.0} & \nb{9.0}\,\dn{58.3} \\
\bottomrule
\end{tabular}
\end{table}

Used as a condition, the same memory already reaches 92.0\% on \taskNoRep, well above the history-conditioned baselines of \cref{tab:suite} (31.7\% and 70.7\%); forming the source from it instead adds 5.0 points on \taskNoRep, consistently across the three seeds, and does not lower the other two tasks (\cref{tab:ablation}).
Removing relocalization costs 12.3 points on \taskNoRep and 11.0 on \taskPair, the two tasks with several items to remember.
Training the anchors through the flow loss alone leaves \taskNoRep unchanged ($+$1.0), but costs 42.4 points on \taskPlate and 58.3 on \taskPair.

\paragraph{Cost.}
\method adds 1.1\,ms per chunk to FM, as memory is read once per chunk by a small module and the transport network sees only the present, whereas the history-conditioned medium- and long-history DP take 37.0 and 34.0\,ms more than \method (\cref{fig:summary}c; \cref{sec:supp-results}).

\subsection{Analysis of the Generation Process}
\label{sec:exp-qual}

\method rests on a division of labor: memory decides where generation starts, and the transport network only completes the motion.
Success rates alone do not show whether the policy works this way, so we examine the generation process itself on \taskNoRep (\cref{fig:qual}).

\paragraph{The choice is made in the source.}
At a decision in which two of the three containers have already been touched, all 30 source samples of \method head to the remaining one before any transport step (\cref{fig:qual}a).
Over 15 decisions, 99.6\% of its source samples head to an untouched container (\cref{fig:qual}b).
In long-history DP, run on the same observations, the choice emerges only during generation: 4.0\% of the samples head to an untouched container at the start, 46.7\% after six of the eight denoising steps, and all of them only at the end.
A history-conditioned policy thus has to settle which target to go to while it generates the motion, the coupling that \method removes.

\paragraph{The transport network only refines.}
The source samples of \method lie a median of 1.7\,cm from the action chunk they become, against 35\,cm for long-history DP.
Generation therefore hardly depends on the number of integration steps: with a single step, \method still reaches 95.0\% on \taskNoRep, against 98.0\% with the default eight (\cref{fig:qual}c), and the same holds on the other tasks tested (\cref{sec:supp-results}).

\subsection{Real-Robot Experiments}
\label{sec:exp-real}

We use an xArm~7 with a parallel gripper, a wrist camera, and a third-person camera, using only their color images (\cref{fig:real}a).
We design four tasks (\cref{fig:real}b) that differ in what must be remembered: a color that was shown, the robot's own earlier actions, or a place marked by a person.
In \emph{Remember Color}, a block of the target color is shown and then hidden under a board, where three blocks are shuffled; once the board is lifted, the robot must pick the block of the shown color.
In \emph{No Repeat}, the real counterpart of \taskNoRep, the robot must touch each of three identical cups exactly once.
In \emph{Marked Plate}, it must place a cube into the plate that a person has set down, and in \emph{Shell Game}, lift the cup under which a person has hidden a block.
We collect 120 demonstrations per task and train \method, DP, FM, and long-history DP for 19{,}600 steps each.
All policies are evaluated at their last checkpoint on the same 24 initial layouts per task, once per layout in a rotated order, with the operator unaware of which policy is running (\cref{sec:supp-real}).
As shown in \cref{fig:real}c, \method achieves the highest success rate on all four tasks, 82.3\% on average against 63.5\% for long-history DP, the strongest baseline; the margin is largest on Remember Color (83.3\% \vs 37.5\%).
Without its memory, \method reduces to FM, which reaches 18.8\% on average.
The memoryless policies never succeed on No Repeat; \cref{fig:real}d illustrates the three behaviors that we observed on this task.

\section{Conclusion}
\label{sec:conclusion}

We presented \method, a flow-policy framework that builds the source distribution from the episodic memory of the robot.
\RR relates the memory to the current view, and \RA turns it into likelihood-trained Gaussian anchors.
On a protocol-matched suite, the public MIKASA-Robo benchmark, and a real robot, this design is most effective when several items must be remembered, and on RoboMimic it does not hurt tasks that need no memory.
Our memory stores visual tokens of single-arm tabletop scenes at a fixed capacity, so tasks with a long and dense relevant past may require a memory that summarizes the past instead of storing its frames.
Extending memory as source to language-conditioned and multi-task policies is a natural next step.

{
    \small
    \bibliographystyle{ieeenat_fullname}
    \bibliography{refs}
}

\clearpage
\maketitlesupplementary
\appendix
\raggedbottom
\renewcommand{\thetable}{S\arabic{table}}
\renewcommand{\thefigure}{S\arabic{figure}}
\setcounter{table}{0}
\setcounter{figure}{0}
\makeatletter\setlength{\@fptop}{0pt}\setlength{\@fpsep}{14pt}\makeatother

\section{Implementation Details}
\label{sec:supp-impl}

\paragraph{Encoder and present tokens.}
Images of $128 \times 128$ are upsampled to $256 \times 256$ and encoded by SigLIP~2-B/16~\cite{siglip2}, fine-tuned end to end with $0.1\times$ the main learning rate.
The $16 \times 16$ patch tokens of each view are average-pooled $2 \times 2$ to $8 \times 8$.
The present $\Zt$ holds the tokens of the current and the previous step from both cameras ($2 \times 2 \times 64 = 256$).

\paragraph{Memory bank.}
After each executed chunk, the $2 \times 64 = 128$ tokens of the chunk's first frame are written with learned camera, position, and age embeddings.
While the memory is empty, a learned null entry takes its place.
The capacity is $C = 32$ entries.
When it is exceeded, for every spatial position the two adjacent entries with the highest cosine similarity are averaged into one, so unchanged positions are merged and changed positions keep their history.

\paragraph{\RR.}
Memory attention has $L_m = 2$ layers with width 256; each layer applies self-attention, cross-attention to $\Mem$, and an MLP, the layer structure of the memory attention in SAM~2~\cite{sam2}.
The slot decoder has $L_q = 2$ layers and $K = 4$ slots.
Heatmaps are normalized per camera over the 64 current-frame tokens of that camera, and $g_k$ concatenates the two pooled features ($2 \times 256$ dimensions), so the anchor MLP receives $[e_k; g_k]$ with 768 dimensions.

\paragraph{\RA.}
$\mathrm{MLP}_a$, $\mathrm{MLP}_\ell$, and $\mathrm{MLP}_\nul$ are two-layer MLPs.
Anchors live in the normalized action space of dimension $H \times d$, with $d = 8$ (seven joint targets and the gripper), and $\sigma_k$ is bounded below by $\sigma_{\min} = 0.05$.

\paragraph{Transport network.}
The transport network is a DiT~\cite{dit} with 16 layers, width 512, and 8 heads, conditioned on $\Zt$ only.
The flow time $\tau \sim \mathcal{U}[0, 1)$ is embedded as $24 \tau$, and inference uses $S = 8$ Euler steps.
Our implementation builds on the public code of GMP~\cite{gmp}, from which we remove the history pathway and replace the denoising objective with rectified flow.

\paragraph{Losses.}
The KL term of \cref{eq:total} is averaged over slots and action dimensions,
\begin{equation}
\mathcal{L}_{\mathrm{KL}} = \frac{1}{KHd} \sum_{k=1}^{K} \mathrm{KL}\big(\N(\mu_k, \sigma_k^2) \,\Vert\, \N(0, I)\big),
\end{equation}
and keeps the anchors from becoming overly narrow.
With $c^\star = \argmax_c r_c$ the component that explains $\Astar$ best, the relaxed winner-takes-all term is
\begin{equation}
\mathcal{L}_{\varepsilon} = -\frac{\varepsilon}{Hd} \sum_{c \neq c^\star,\, c \neq \nul} \log \N(\Astar; \mu_c, \sigma_c^2),
\end{equation}
with $\varepsilon = 0.05$.
It applies the likelihood with a small weight to the anchors that do not explain $\Astar$ best, so that no slot is left untrained.
We use $\beta = 10^{-3}$ and $\lambda = 1$.

\paragraph{Training.}
We use AdamW with $\beta = (0.95, 0.999)$ and weight decay $10^{-6}$, and a cosine learning-rate schedule that peaks at $3 \times 10^{-4}$ after 2{,}000 warm-up steps.
During the first 2{,}000 steps, the transport network is trained from $\N(0, I)$ and the anchors only by likelihood.
The image encoder receives gradients only from the flow loss, as in FM: \RR and the no-recall logit read stop-gradient encoder tokens, both present and remembered.
During training, the memory bank is rolled out chunk by chunk with the same code as at inference.

\paragraph{Inference.}
In stochastic mode, a component $c$ is sampled from $w$, and $x_0 \sim \N(\mu_c, \sigma_c^2)$, where the no-recall component is $\N(0, I)$; the deterministic mode is given in \cref{sec:train}.
Every chunk is executed open loop in full ($H$ steps).

\paragraph{Implementation checks.}
We verified that: with an empty memory the source is exactly the no-recall component; with memory disabled, the forward pass is bitwise identical to FM; anchors at chunk $j$ do not change when later chunks are modified; and memories and anchors computed during training equal those computed at inference.
Each check fails when the corresponding property is deliberately broken.

\begin{table}[h]
\centering
\caption{Hyperparameters of \method, shared across benchmarks unless noted.}
\label{tab:supp-hparams}
\small
\begin{tabular}{@{}>{\raggedright\arraybackslash}p{0.52\linewidth}>{\raggedright\arraybackslash}p{0.40\linewidth}@{}}
\toprule
Parameter & Value \\
\midrule
Recall slots $K$ & 4 \\
Memory width / capacity $C$ & 256 / 32 entries \\
Memory-attention / slot-decoder layers & 2 / 2 \\
$\sigma_{\min}$ & 0.05 \\
KL weight $\beta$, relaxed WTA $\varepsilon$, $\lambda$ & $10^{-3}$, 0.05, 1 \\
Warm-up & 2{,}000 steps \\
Euler steps $S$ & 8 \\
Chunk length $H$ & 8 (\suite); on MIKASA-Robo 10 (RC), 8 (IM, SGT) \\
\bottomrule
\end{tabular}
\end{table}

\section{\suite Details}
\label{sec:supp-suite}

\cref{tab:supp-suite} summarizes the tasks.
Scenes are built on ManiSkill3~\cite{maniskill3} with the MIKASA-Robo framework~\cite{mikasa}; the third-person and wrist cameras and the robot are those of MIKASA-Robo.
\taskPlate and \taskNoRep are related to the real-robot tasks of~\cite{chameleon}, and \taskCount to the SwingXTimes task of RoboMME~\cite{robomme}.
The code of \suite and \method will be released.

\begin{table*}[t]
\centering
\caption{\textbf{Tasks of \suite.} We list what each task requires the robot to remember, its candidates, and the chance level: the success rate of a policy that executes every motion perfectly but guesses what must be remembered uniformly at random (\cref{sec:supp-suite}).}
\label{tab:supp-suite}
\small
\setlength{\tabcolsep}{5pt}
\begin{tabular}{@{}llcc@{}}
\toprule
Task & What must be remembered & Candidates & Chance (\%) \\
\midrule
\taskPlate & which plate the marker touched & 5 plates & 20.0 \\
\taskCount & how many swings were already made & stop or swing again & 11.7 \\
\taskNoRep & which containers were already touched & 3 containers & 22.2 \\
\taskNoRepFour & which containers were already touched & 4 containers & 9.4 \\
\taskPair & which two containers the marker hovered over & 4 containers & 16.7 \\
\bottomrule
\end{tabular}
\end{table*}

\paragraph{Tasks.}
In \taskPlate, a marker without collision geometry descends into one of five identical plates, 12\,cm apart, pauses, and rises away; the robot holds a cube from the start and must place it into the marked plate.
In \taskNoRep, the robot must touch each of three identical containers, 15\,cm apart, exactly once, in any order; touching leaves no visible trace.
\taskNoRepFour adds a fourth container at the same spacing.
\taskPair uses the scene of \taskNoRepFour: a marker without collision geometry hovers above two of the four containers in turn and then disappears; which two are cued, and in which order, is drawn uniformly.
The robot stays still while the marker is shown and must then touch each cued container once, in any order.
In \taskCount, a row of five indicator slots in front of the robot shows $N \in \{2, 3, 4, 5\}$, drawn uniformly, as $N$ red discs followed by gray ones, lit throughout the episode; the robot starts with the cube in its gripper, must swing it between a right and a left pad $N$ times, and then put it on a stop plate.
A swing counts when the held cube enters the zone above the right pad and then the zone above the left pad, within 3\,cm of the pad center horizontally and below 12\,cm; the whole layout is jittered by up to 2\,cm.

\paragraph{Demonstrations.}
Each task has 250 demonstrations generated by a motion-planning script under joint-position control, the same action space as at evaluation.
Hidden answers are balanced across demonstrations.
While a cue or event is shown, the expert holds a fixed pose, so that the arm configuration does not encode the answer.
In \taskCount, the expert first waits for 0, 24, or 48 steps, drawn at random, and its joint angles in the frame before each decision agree to within $7.6 \times 10^{-6}$\,rad.

\paragraph{Success and chance.}
An episode succeeds only if the first choice is correct and the task is completed.
For \taskNoRep and \taskNoRepFour, the choice is correct if no container is touched twice and none is skipped.
For \taskPair, the task is completed once both cued containers have been touched, and the choice is correct if by then no other container has been touched and none has been touched twice.
For \taskCount, the choice is correct if exactly $N$ swings have been made when the cube is first released, and the task is completed if the cube lands on the stop plate.
The chance level of \cref{tab:suite} is the success rate of a policy that executes every motion perfectly but guesses what must be remembered uniformly at random: the plate in \taskPlate ($1/5$); after each swing, whether to swing again in \taskCount ($2^{-N}$, averaged over $N$); the container at each touch in \taskNoRep and \taskNoRepFour ($3!/3^3$ and $4!/4^4$); and the cued pair in \taskPair ($1/\binom{4}{2}$).

\paragraph{Validation.}
On \taskPlate, \taskNoRep, and \taskCount, the scripted expert, replayed with the evaluation seeds, succeeds in all 100 evaluation episodes; \taskPair and \taskNoRepFour reuse the scene, the expert, and the success logic of \taskNoRep.
At the decision moment, a CNN that sees only the current frame and an MLP on the joint angles predict the answer at chance: 19.5\% and 17.0\% on \taskPlate (chance 20.0\%), and a balanced accuracy of 47.6\% and 47.8\% on \taskNoRep, where they predict for each container whether it has already been touched (chance 50.0\%).
Trained on a frame in which the answer is visible, the same CNN reaches 100\%, so the test detects the answer when it is in view.
On \taskCount, the two probes cannot tell whether to stop (ROC-AUC 0.501 and 0.493), while the same CNN reads $N$ from the indicators with 100\% accuracy.
The logic that records touches and placements in \taskPlate and \taskNoRep has unit tests, each of which fails when the logic is deliberately broken.

\section{Baselines}
\label{sec:supp-baselines}

\cref{tab:supp-baselines} lists the baselines.
All DP variants run the diffusion-policy implementation of the codebase~\cite{gmp} with its released configuration, which uses the same image encoder as \method: SigLIP~2-B/16, fine-tuned with $0.1\times$ the main learning rate.
Their denoising network is a conditional transformer of the same size as the transport network of \method (16 layers, width 512, 8 heads), and they sample with DDIM in 8 steps, as many as the Euler steps of \method.
Medium- and long-history DP are the released history configurations of~\cite{gmp}, conditioned on the frames of the past 16 and 120 steps.
Every decision in our tasks looks back at most 120 steps (at most 106 on \taskNoRepFour and 77 on \taskPair, and 67, 73, 116, and 81 on the four real-robot tasks), so the long-history window covers every cue.\pullup

\begin{table*}[t]
\centering
\caption{\textbf{Baseline configurations.} All policies that we train on a benchmark share demonstrations, cameras, the number of training steps on each task, and evaluation episodes. The DP variants use the image encoder of \method and a denoising network of the same size as its transport network (\cref{sec:supp-baselines}).}
\label{tab:supp-baselines}
\small
\setlength{\tabcolsep}{5pt}
\begin{tabular}{@{}llll@{}}
\toprule
Method & Implementation & History & Generation \\
\midrule
DP~\cite{dp} & code of~\cite{gmp}, no history & current step & DDIM, from $\N(0, I)$ \\
FM & ours, memory disabled & current and previous step & rectified flow, from $\N(0, I)$ \\
Medium-history DP & code of~\cite{gmp} & past 16 steps & DDIM, from $\N(0, I)$ \\
Long-history DP & code of~\cite{gmp} & past 120 steps, one frame per 8 steps & DDIM, from $\N(0, I)$ \\
\method & ours & episodic memory, through the source only & rectified flow, from $p_0$ \\
\bottomrule
\end{tabular}
\end{table*}

\paragraph{Training budget.}
Within a benchmark, all policies that we train use the same number of gradient steps on each task.
On \suite, \taskPlate and \taskCount are trained for 19{,}600 steps, and \taskNoRep, \taskNoRepFour, and \taskPair for 39{,}200 steps.

\section{MIKASA-Robo Protocol}
\label{sec:supp-mikasa}

As in GMP~\cite{gmp}, we train one policy per task.
We use the first 250 demonstrations of each task in the public dataset~\cite{mikasa}, converted to absolute joint targets as in~\cite{gmp}, and train for up to 51 epochs with the optimizer of \cref{sec:supp-impl}; the chunk lengths are listed in \cref{tab:supp-hparams}.
Every checkpoint is evaluated on the same 100 episodes in deterministic mode with the official success criterion of MIKASA-Robo, and, like GMP~\cite{gmp}, we report the best of the periodically evaluated checkpoints.

\paragraph{Two-task comparison.}
\cref{tab:supp-mikasa-two} compares \method with the methods that are evaluated only on ShellGameTouch and RememberColor3~\cite{vpwem}.

\begin{table}[t]
\centering
\caption{\textbf{Comparison on ShellGameTouch and RememberColor3} of MIKASA-Robo, the two tasks on which several methods are evaluated~\cite{vpwem}. We report the success rate (\%); some sources give only the average.}
\label{tab:supp-mikasa-two}
\small
\setlength{\tabcolsep}{6pt}
\begin{tabular}{@{}lccc@{}}
\toprule
Method & SGT & RC3 & \textbf{Avg.}\,$\uparrow$ \\
\midrule
DP~\cite{dp,vpwem} & \na & \na & 19.5 \\
DP-PTP~\cite{ptp,vpwem} & \na & \na & 15.0 \\
MaIL~\cite{mail,vpwem} & \na & \na & 19.5 \\
BPP~\cite{bpp,vpwem} & 94 & 21 & 57.5 \\
MemER~\cite{memer,vpwem} & 82 & 27 & 54.5 \\
DP-VPWEM~\cite{vpwem} & 91 & 82 & 86.5 \\
Chameleon~\cite{chameleon} & \sbest{96.7} & \sbest{94.4} & \sbest{95.6} \\
\oursrow \textbf{\method} & \textbf{100.0} & \textbf{100.0} & \textbf{100.0} \\
\bottomrule
\end{tabular}
\end{table}

\section{RoboMimic Protocol}
\label{sec:supp-robomimic}

We use Square with proficient (ph) and multi-human (mh) demonstrations.
On this benchmark, policies observe $256 \times 256$ images together with the 10-dimensional proprioceptive state of the current step.
DP, FM, and \method are trained with the same number of gradient steps, and every checkpoint is evaluated on 100 episodes.
\cref{tab:robomimic} lists the success rates behind \cref{fig:summary}b; its long-history DP row is the result reported in~\cite{gmp} for this task.\pullup

\begin{table}[t]
\centering
\caption{\textbf{Comparison on RoboMimic~\cite{robomimic}} behind \cref{fig:summary}b. We report the success rate (\%) on Square with proficient (ph) and multi-human (mh) demonstrations, averaged over three seeds. $^\dagger$Reported in~\cite{gmp}.}
\label{tab:robomimic}
\small
\setlength{\tabcolsep}{8pt}
\begin{tabular}{lccc}
\toprule
& \multicolumn{2}{c}{Square} & \\
\cmidrule(lr){2-3}
Method & ph & mh & \textbf{Avg.}\,$\uparrow$ \\
\midrule
\grprow{4}{Memoryless}
DP~\cite{dp} & \textbf{98.0} & 73.3 & 85.7 \\
FM & 97.0 & 75.3 & 86.2 \\
\midrule
\grprow{4}{History-conditioned}
Long-history DP$^\dagger$ & 86 & 69 & 77.5 \\
\midrule
\grprow{4}{Memory as source}
\oursrow \textbf{\method} & 94.0 & \textbf{84.0} & \textbf{89.0} \\
\bottomrule
\end{tabular}
\end{table}

\section{Real-Robot Setup}
\label{sec:supp-real}

The robot is an xArm~7 with the xArm gripper, a wrist-mounted RGB-D camera, and a third-person RGB-D camera; only color images at $128 \times 128$ are used.
We collect 120 demonstrations per task, with the hidden answer balanced.
Each of the four policies is trained for 19{,}600 steps and evaluated at its last checkpoint on the same 24 initial layouts per task.
Every layout is evaluated once with each policy, in an order rotated by a Latin square, and the operator does not know which policy is running.
A trial ends when the robot has stopped moving for 5\,s (on No Repeat, when it has returned to its initial pose for 2\,s) or at a time limit of 30\,s (45\,s on No Repeat).
Success is judged from the recordings of both cameras with the criterion of \suite: the first choice is correct and the task is completed.
Policies are identified only by letters during evaluation, and all trials are judged before the letters are revealed.
An operator with an emergency stop is present during all robot motion.
\cref{tab:supp-real} lists the successful trials of each policy on each task, which \cref{fig:real} shows as rates.

\begin{table}[t]
\centering
\caption{\textbf{Real-robot results} behind \cref{fig:real}. We report the number of successful trials out of 24 per task and the success rate (\%) over all 96 trials.}
\label{tab:supp-real}
\small
\setlength{\tabcolsep}{3.2pt}
\begin{tabular}{@{}lccccc@{}}
\toprule
Method & \begin{tabular}[b]{@{}c@{}}Remember\\ Color\end{tabular} & \begin{tabular}[b]{@{}c@{}}No\\ Repeat\end{tabular} & \begin{tabular}[b]{@{}c@{}}Marked\\ Plate\end{tabular} & \begin{tabular}[b]{@{}c@{}}Shell\\ Game\end{tabular} & Avg. \\
\midrule
DP~\cite{dp} & 9 & 0 & 12 & 7 & 29.2 \\
FM & 5 & 0 & 8 & 5 & 18.8 \\
Long-history DP & 9 & 14 & 22 & 16 & 63.5 \\
\method & \textbf{20} & \textbf{17} & \textbf{24} & \textbf{18} & \textbf{82.3} \\
\bottomrule
\end{tabular}
\end{table}

\paragraph{Tasks.}
\emph{Remember Color}: a block of the target color is shown and covered by a board, three blocks including one of that color are shuffled under the board, and once the board is lifted the robot picks the block of the shown color.
\emph{No Repeat}: three identical cups stand upside down, and the robot must touch the bottom of each cup exactly once; touching a cup twice or skipping one is a failure. It is the real counterpart of \taskNoRep.
\emph{Marked Plate}: a person brings one of three identical plates and sets it down; the robot then picks up a cube and places it into that plate.
\emph{Shell Game}: three identical opaque cups stand upside down; a person puts a small cube under one of them and covers it again, and once the hand has left, the robot must grasp the cup that hides the cube and lift it to reveal the cube.
Related tasks in prior work are RememberColor and ShellGame in MIKASA-Robo~\cite{mikasa}, whose simulated ShellGameTouch is used in \cref{tab:mikasa}, the seasoning, plate, and cup tasks of~\cite{chameleon}, and TrackCube in RoboMME~\cite{robomme}.

\section{Ablation Details}
\label{sec:supp-ablation}

\cref{tab:supp-ablation-cue} reports the variants of \cref{tab:ablation} on \taskCount, trained with one seed and the budget of the task (\cref{sec:supp-baselines}).
On this single-item task, every variant that keeps the memory succeeds in at least 92\% of the episodes, in line with \cref{tab:suite}, where long-history DP also solves the single-item tasks.

\begin{table}[t]
\centering
\caption{\textbf{Ablation study on \taskCount.} We report the success rate (\%) of one training seed for the variants of \cref{tab:ablation}.}
\label{tab:supp-ablation-cue}
\small
\setlength{\tabcolsep}{8pt}
\begin{tabular}{@{}lc@{}}
\toprule
Variant & \taskCount \\
\midrule
\method & 98.0 \\
\midrule
W/o memory & 6.0 \\
As condition & 100.0 \\
W/o relocalization & 92.0 \\
Flow loss only & 97.0 \\
\bottomrule
\end{tabular}
\end{table}

\section{Additional Results}
\label{sec:supp-results}

\paragraph{Training seeds.}
Averaged over the five tasks of \cref{tab:suite}, the three training seeds give $6.0 \pm 0.3$ for DP, $7.6 \pm 1.4$ for FM, $29.7 \pm 0.4$ for medium-history DP, $64.3 \pm 4.6$ for long-history DP, and $84.6 \pm 0.4$ for \method (mean $\pm$ standard deviation, as in \cref{tab:mikasa}); averaged over Square (ph) and (mh) in \cref{fig:summary}b, they give $85.7 \pm 1.4$ for DP, $86.2 \pm 0.6$ for FM, and $89.0 \pm 1.3$ for \method.

\paragraph{Fewer integration steps.}
\cref{tab:supp-steps} evaluates the seed-0 checkpoints of \method with 1, 2, 4, and 8 Euler steps on the same episodes; \cref{fig:qual}c shows the same evaluation on \taskNoRep.

\begin{table}[t]
\centering
\caption{\textbf{Fewer integration steps.} We evaluate the seed-0 checkpoints of \method with 1, 2, 4, and 8 Euler steps (the default is 8) and report the success rate (\%). The 8-step entry of \taskCount differs by one episode from its seed-0 entry in \cref{tab:supp-ablation-cue}, as this analysis was run on a different machine.}
\label{tab:supp-steps}
\small
\setlength{\tabcolsep}{6pt}
\begin{tabular}{@{}lcccc@{}}
\toprule
Steps & 1 & 2 & 4 & 8 \\
\midrule
\taskPlate & 97.0 & 97.0 & 95.0 & 95.0 \\
\taskCount & 98.0 & 100.0 & 99.0 & 99.0 \\
\taskNoRep & 95.0 & 96.0 & 96.0 & 98.0 \\
\bottomrule
\end{tabular}
\end{table}

\paragraph{Memory capacity.}
Evaluating the same checkpoints with half the memory capacity, $C = 16$, so that entries are merged 8 times per episode on \taskNoRepFour and 3 times on \taskPair, leaves the success rate of every seed unchanged.\pullup

\paragraph{Cost.}
\cref{tab:supp-cost} reports the inference cost behind \cref{fig:summary}c.

\begin{table}[t]
\centering
\caption{\textbf{Inference cost} behind \cref{fig:summary}c. We measure each policy on one RTX 4090 with batch size 1 and report the median latency per action chunk over 34 chunks, its increase over FM, and the peak GPU memory.}
\label{tab:supp-cost}
\small
\setlength{\tabcolsep}{6pt}
\begin{tabular}{@{}llr@{}}
\toprule
Method & Latency (ms)\,$\downarrow$ & Memory (GB)\,$\downarrow$ \\
\midrule
DP~\cite{dp} & 148.1 & 1.10 \\
FM & 152.6 & 1.10 \\
Medium-history DP & 190.7\,\dl{+38.1} & 1.60 \\
Long-history DP & 187.7\,\dl{+35.1} & 1.60 \\
\oursrow \textbf{\method} & 153.7\,\dl{+1.1} & 1.14 \\
\bottomrule
\end{tabular}
\end{table}

\end{document}